\documentclass{article}

\usepackage{arxiv}

\usepackage[utf8]{inputenc} 
\usepackage[T1]{fontenc}    
\usepackage{hyperref}       
\usepackage{url}            
\usepackage{booktabs}       
\usepackage{amsfonts}       
\usepackage{nicefrac}       
\usepackage{microtype}      
\usepackage{lipsum}
\usepackage{graphicx}
\graphicspath{ {./images/} }

\usepackage[utf8]{inputenc}
\usepackage[english]{babel}
\usepackage[LAE,T1]{fontenc}

\title{OSN-MDAD: Machine Translation Dataset for Arabic Multi-Dialectal Conversations on Online Social Media.}

\author{
 Fatimah Alzamzami \\
  Multimedia Communication Research Laboratory\\
  School of Electrical Engineering and Computer Science\\
  University of Ottawa, Canada\\
  \texttt{falza094@uottawa.ca} \\
   \And
 Abdulmotaleb El Saddik \\
  Multimedia Communication Research Laboratory\\
  School of Electrical Engineering and Computer Science\\
  University of Ottawa, Canada\\
  Mohamed bin Zayed University of Artifcial Intelligence\\
  Abu Dhabi, UAE\\
  \texttt{elsaddik@uottawa.ca} \\
}

\begin{document}
\maketitle
\begin{abstract}
While the resources for English language are fairly sufficient to understand and analyze content on social media, similar resources in other languages, such as Arabic, are still immature. The main reason that the resources in Arabic are insufficient is that Arabic has many dialects in addition to the Modern Standard Arabic (MSA). Arabic speakers do not use the standard version (MSA) in their day-to-day communications and conversations; rather, they use dialectal versions. Unfortunately, social users transfer this phenomenon (i.e. language preference) into their use of digital social media platforms, which in turn has raised an urgent need for and challenges to building suitable datasets and AI models for different systems and applications. Existing machine translation systems designed for MSA fail to work well with Arabic dialects. In light of this, it is necessary to adapt to the informal nature of communication on online social networks (OSNs) by developing translation systems that can effectively handle the various dialects of Arabic. 
Unlike for MSA that shows advanced progress in translation systems, little effort has been exerted to utilize Arabic dialects for machine translation systems. While a few attempts have been made to build translation datasets for dialectal Arabic, they are domain dependent and are not OSN cultural language friendly.
In this work, we attempt to alleviate these limitations by proposing an online social network-based multidialect Arabic dataset (OSN-MDAD) that is crafted by contextually translating English tweets into four Arabic dialects: Gulf, Yemeni, Iraqi, and Levantine/Shami. To perform the tweet translation, we followed our proposed guideline framework for content translation, which could be universally applicable for translation between foreign languages and local dialects. We validated the authenticity of our proposed OSN-MDAD dataset by developing neural machine translation (NMT) models for four Arabic dialects, one for each dialect. Transfer learning and Transformer's architecture were adopted to build the four translation models. Our results have shown a superior performance of the Transformer-based models (F-score of BLUE and ROUGE $\ge$ 30 scores) compared to the performance of the sequence-to-sequence with attention mechanism (F-score of BLUE and ROUGE $\le$ 13.1 scores) when training using OSN-MDAD dataset. We believe that our proposed dataset can serve as a benchmark Arabic multidialectal translation dataset for machine translation tasks, as the experimental results have shown.
\end{abstract}


\section{Introduction}
\label{sec:introduction}

In today’s digital and global age, $\approx$ 52\% of internet users access and post information in their own language and dialect. Seventy-five percent of world’s population do not speak English, and French, Arabic and Spanish are among the most spoken languages after English on social media. In order to accurately analyze social behaviors of social users who speak different languages, we need to understand the expressions of the languages they use to share and communicate their opinions and thoughts \cite{ArSarcasm}. One current approach to addressing the
multilingual aspect on OSNs is to use a translation approach between languages \cite{balahur2012multilingual, dashtipour2016multilingual, 9430413}. The challenge with the translation approach is that it heavily depends on the quality of the parallel translation datasets \cite{tran2018improving, alam2022roman, mohamed2023alphamwe}. Low-quality datasets lead to low-quality models and vice versa \cite{tran2018improving, dashtipour2016multilingual, alam2022roman, tomalin2021practical}. Hence, a careful content translation methodology need to be considered when crafting translation datasets.

Th advent of social media has revolutionized the language that people use for communication; the informal nature of conversations and communication has become the new norm on online social media platforms like Twitter. This is plainly evident in the Arabic language, where urban dialects have become the dominant means of communication instead of Modern Standard Arabic (MSA), which is different from dialectal Arabic in terms of morphology, lexicons, and expressions. As a result, existing translation systems designed for MSA fail to work well with Arabic dialects \cite{gamal2022survey, babaali2022survey}. In light of this, it is necessary to adapt to the informal nature of communication on OSNs by developing translation systems that can effectively handle the various Arabic dialects in addition to MSA. Unlike MSA, for which considerable progress has been made in translation systems such as Google Translator, little effort has been exerted to include Arabic dialects in machine translation systems. The main limitation contributing to the immaturity of Arabic dialect translation systems is the insufficiency of data resources and datasets \cite{babaali2022survey}. Only few efforts \cite{kchaou2023hybrid} have been made to build datasets to support the Arabic multidialect translation systems ; however, the construction design of those dataset is not OSN-cultural language friendly, which might limit the translation performance of Arabic multidialect systems on OSNs. The nature of conversations on OSNs involves short and unstructured messages. In addition to hashtags and short forms, slang expressions are commonly used on OSNs. With the informal language of day-to-day talks being transferred on OSNs, idiomatic expressions such as "it is a piece of cake" and "it is raining cats and dogs" are naturally used within messages. This has resulted in a new nature of contents with a mix of spoken and online language that the existing datasets have ignored. Therefore, this linguistic representational challenge must be be considered to utilize the online language in machine translation systems. It is well known that OSNs platforms (Twitter in our work case) are an open environment for conversations of different topics and domains; hence, domain-specific data resources (The Bible\footnote{\label{bible1} https://www.biblesociety.ma}, \footnote{\label{bible2} https://www.bible.com}, \cite{takezawa2007multilingual}) introduce a major limitation to the capability of translation models \cite{zakraoui2021arabic}. That being said, the domain-specific approach in designing translation datasets narrows the spectrum of vocabularies and expressions to be solely specific to the particular domains. This limitation, in turn, has raised other challenges, such as out-of-vocabulary (OOV) and out-of-domain test data, which therefore would affect the generalizability or adaptability of such datasets to open contents existing on OSNs.

In addition to the insufficiency of Arabic multidialectal data resources \cite{kchaou2023hybrid}, the protocol followed for dataset crafting and content translation in the existing datasets is not clear. 
The criteria and guidelines for choosing source sentences, translating content, qualifying translators, and evaluating the quality of target translations are ambiguous and are sometimes absent \cite{bouamor2014multidialectal, zbib2012machine, bouamor2018madar}. In this paper, we aim to overcome the abovementioned challenges and propose an OSN-based multidialect Arabic dataset (OSN-MDAD). OSN-MDAD is created by contextually translating English tweets into four main urban dialects: Gulf, Yemeni, Iraqi, and Levantine/Shami. We do not accept word-to-word translation; instead, we propose a contextual translation of tweets from English to multidialect Arabic. We adopt the content localization approach in our translation. The proposed dataset is designed and built to work in the wild under the uncontrolled conditions of content sharing on social media. To the best of our knowledge, we are the first to construct a multidialectal Arabic-English parallel dataset for machine translation that is optimized for the informal language used on social media. Moreover, we are the first to create a guideline framework that is applicable for translating not only a foreign language into different Arabic dialects , but also from any language to any other language and dialect.

We summarize the contributions of this work as follows:

\begin{itemize}

\item Design a guideline framework for constructing online social media (OSN)-based contextual translation datasets.
\item Construct a benchmark translation dataset by translating tweets from English into multidialect Arabic (i.e., Gulf, Levantine/Shami, Iraqi, and Yemeni).
\item Develop four OSN-based English-to-Arabic deep learning-based translation models for four Arabic dialects: Gulf, Levantine/Shami, Iraqi, and Yemeni.

\end{itemize}

The rest of the paper is organized as follows. Section \ref{sec:rw}
presents in detail the related works in the literature and our proposed methodology is presented in Section \ref{sec:methodology}. Section \ref{sec:exprResult} explains the experimental design and evaluation protocol followed in this work followed by an illustration and discussion of the results and analysis.
Finally, in Section \ref{sec:conclusion} we summarize our proposed work and findings and discuss future directions.
\section{Related Work}
\label{sec:rw}

The phenomenon of urban dialects (i.e. informal conversations) has become the dominant mean of communication on online social media platforms such as Twitter. This phenomenon is plainly evident in the Arabic language, where the informal dialects have become more common than Modern Standard Arabic (MSA), for communication among social users on OSNs. Modern Standard Arabic (MSA) is different from the dialectal Arabic in terms of morphology, lexicons, and expressions.
As a result, existing translation systems designed for MSA fail to work well with Arabic dialects \cite{emna2022neural}. Considering this, it is necessary to adapt to the informal nature of communication on OSNs by developing translation systems that can effectively handle the multidialect aspect of Arabic language, besides the MSA. Unlike MSA that shows advanced progress in translation systems such as Google Translator, little efforts have been exerted to utilize Arabic dialects for machine translation systems \cite{kchaou2023hybrid}. The main limitation contributing to the immaturity of Arabic dialect translation systems is the insufficiency of data resources and datasets \cite{emna2022neural}. Only a few efforts have been made to create datasets for dialectal Arabic. The Bible was among the first data resources that was translated into Arabic Moroccan \footref{bible1} and Arabic Tunisian dialects \footref{bible2}.  In the context of online data, Zbib et al. \cite{zbib2012machine} created Egyptian-to-English and Levantine-to-English dataset collected from dialectal Arabic weblogs and online user groups \footnote{https://catalog.ldc.upenn.edu/LDC2012T09}. The translation from Arabic to English was carried out through crowdsourcing technique on Mechanical Turk by Arabic users. Bouamor et al. \cite{bouamor2014multidialectal} used a subset of 2K sentences from Zbib’s dataset and extended the translation to Palestinian, Syrian, and Tunisian dialects. The authors asked native dialectal Arabic speakers to translate  Egyptian sentences into their own native dialects. Later, Bouamor et al. \cite{bouamor2018madar} created parallel-phrase and parallel-sentence datasets that cover various city-level Arabic dialects. The corpus was created by translating a subset of phrases and sentences taken from the Basic Traveling Expression Corpus (BTEC) \cite{takezawa2007multilingual}. Although those Arabic dialect datasets were created as an attempt to support the Arabic multidialect translation systems, they reveal a number of issues that might limit the translation performance of Arabic multidialect systems on online social networks (OSNs). Nine observed reasons support this claim: (1) the number of sentences translated into each dialect is small; some datasets comprise as few as 2K sentences per dialect  \cite{bouamor2018madar}; (2) translation was done by non-professional translators, and it is unknown whether the translators were native speakers of an Arabic dialect and possessed sufficient/native English proficiency to ensure accurate translation \cite{zbib2012machine}; (3) translators were not screened regarding whether they were familiar with informal and slang English and with OSN language (i.e., whether they were active on social media); (4) some datasets are domain dependent \footref{bible1}, \footref{bible2},  \cite{takezawa2007multilingual, nassif2021empirical} and might not be adapted to other domains; (5) idiomatic expressions were not taken into consideration; (6) most of the datasets did not consider identical sentence translation into different dialects; (7) ONS cultural expressions were not included in the datasets and were not translated; (8) code-borrowing terms were not taken into consideration; and (9) the translation criteria and guidelines for most of the datasets have some ambiguity or are even absent \cite{bouamor2014multidialectal, zbib2012machine, bouamor2018madar}. 
To overcome the above limitations and to complement the existing datasets of dialectal Arabic  for machine translation on social media, we propose an OSN-based multidialect Arabic dataset (OSN-MDAD). OSN-MDAD is created by contextually translating English tweets into four main urban dialects: Gulf, Yemeni, Iraqi, and Levantine/Shami. To the best of our knowledge, we are the first to construct a multidialectal Arabic-English dataset for machine translation that is optimized for the informal language used on social media. Moreover, we are the first to create a guideline framework that is applicable for translating not only a foreign language into different Arabic dialects but also for translating any language to any other language and dialect.

Machine translation models trained on MSA data are not capable of performing well in translating from and into Arabic dialects \cite{sajjad2020arabench}. Hence, research efforts have been made to enrich such models to explicitly handle the translation of Arabic dialects. The very first efforts toward building an Arabic dialectal machine translation were based on statistical learning approach \cite{salloum2011dialectal, sajjad2016egyptian}. Some studies adopted the approach of appending a dialect to MSA as a preprocessing step \cite{salloum2011dialectal,salloum2011dialectal,salloum2013dialectal}. Other studies adapted MSA-to-English systems to dialectal data \cite{sajjad2016egyptian}. Recently, deep learning methods have become more dominant than statistical learning approaches, and neural machine translation (NMT) has been shown to outperform STM with state-of-the-art results \cite{costa2018neural}. NMT for Arabic dialects has not been extensively explored; however, some work on translating Arabic dialects using NMT has been recently introduced \cite{baniata2022reverse, slim2022improving, gamal2022survey, kchaou2023hybrid}. Baniata et al. \cite{baniata2018neural} designed a recurrent neural network-based encoder-decoder neural machine translation model that uses multitask learning with individual encoders for both MSA and dialects and with a shared decoder. Transformer neural networks have become an alternative to sequence-to-sequence neural networks for neural machine translation. Shapiro and Duh \cite{shapiro2019comparing} trained transformer-based models for MSA, Egyptian, and Levantine dialects. Their results showed that training a model on multidialectal data can benefit the translation of unknown dialects. Sajjad et. al. \cite{sajjad2020arabench} found that transformer based NMT models produced better translations when trained on large-scale datasets compared to small-sized datasets in the settings of training from scratch. On the other hand, their results revealed that using transfer learning through fine-tuning pretrained NMT models (i.e. on MSA) improved the learning of Arabic dialectal translations. In our work, we use transformer networks and transfer learning to train our NMT models using our proposed OSN-MDAD dataset.  
\section{Methodology}
\label{sec:methodology}

\subsection{OSN-based English to Multi-dialect Arabic Translation 
 Dataset}
 
 Our dataset is designed to offer high quality translation models for content on online social networks (OSNs), where people mostly use informal and slang language to express feelings, opinions, and thoughts. The goal of this dataset is to offer a high-quality resource to translate informal and slang English into informal and slang multidialect Arabic. Our translation strategy is based on the content localization approach and not word-to-word translation. The content localization translation approach is more precise since it translates content to be culturally relevant and easy to understand in context.  In addition, idiomatic expressions are considered when localizing our English dataset. Moreover, quality control is applied to monitor the translators’ work and ensure the quality of the translated datasets. To guarantee the quality of the final translations, we carefully decided on three main aspects: translators, data subsets, and translation strategy.
 
\subsubsection{Translators}
 We required that all participants be professional translators, native in the Arabic dialects, and native or fluent in English, and familiar with informal and slang English. Additionally, the translators are required to have social involvement on social media (i.e. Twitter, Facebook, YouTube or any other platform) using both English and native Arabic dialects. Furthermore, female and male translators were involved in our translation task to minimize gender bias in translation; many studies \cite{fujita1991gender, wierzbicka1999emotions, tomalin2021practical} reflect that the way people perceive others' opinions and express their own are subject to gender differences.

Initially, we wanted to use crowdsourcing platforms to perform our translation, but there were not enough native dialect Arabic workers available, and for some dialects, there were zero workers. Alternatively, we asked professional translators who are native speakers of Arabic dialects to conduct the translation task. An invitation email was sent to a list of professional translators for the Gulf, Iraqi, Yemeni, and Levantine/Shami dialects. The invitation included a questionnaire asking for their personal and demographic information as well as their social involvement on social media. Prior to accepting a translator, we first conducted individual audio calls with potential translators to ensure that they were native speakers in their Arabic dialect. Then each candidate underwent a 2-stage qualification test as follows:

\begin{enumerate}

\item Qualification Test Stage 1:

	\begin{enumerate}
	\item We provided the translator with a list of English test tweets that had been carefully chosen to include slang expressions, social medial language expressions, and idiomatic expressions. We provided the translator with our translation guidelines and asked them to localize the tweets in their native Arabic dialect while preserving the context, tone, and content. 
	
	\item A corresponding native Arabic dialect speaker checked the translation to approve or reject it.  
	
	\end{enumerate}
	
\item Qualification Test Stage 2:

	\begin{enumerate}
	\item If the translator passed qualification test stage 1, we provided the first tranche of tweets (i.e., 500 tweets) to be translated, and then we checked 50\% of the translated tweets randomly. 
	
	\item A corresponding native Arabic dialect speaker checked the translations to approve or reject them. If the translator passed the qualification test stage 2, he or she was qualified as a translator in our translation task. 
	
	\end{enumerate}	

\end{enumerate}

According to our criteria, 13 out of 39 translators were qualified to participate in our translation task: three translators each for the Gulf, Yemeni, and Iraqi dialects and four translators for the Levantine/Shami dialect.  Fifty-four percent of the participants were females, and 46\% were males. All the candidates were native speakers of their spoken Arabic dialect, native or fluent in English, familiar with informal and slang English, and had social presence and engagements on social media.

\subsubsection{Tweet Subsets}
 The English tweets to be translated were taken from our DFMSD dataset \cite{abaalkhail2018affectional}, and $\approx$ 500 sentences with idiomatic expressions were appended to the tweets list, resulting in a total size of 15000 sentences. 
 
DFMSD (Domain Free Sentiment Multimedia Dataset) \cite{abaalkhail2018affectional} is an English tweet dataset that followed high standard criteria for the data collection and preparation procedures. DFSMD was collected whiteout a restriction of filters such as keywords, geo-locations, topics, location, etc., in order to respect the purpose of creating a generalized (i.e. domain-free) dataset independent of events, topics, users, and emotions. The tweets were collected worldwide from different times and dates, chosen randomly to ensure that as many of the topics and events discussed daily were covered. To ensure that the content of all tweets was appropriate and distinct, we filtered out retweets, tweets with inappropriate content, and  non-useful content (i.e. tweets with only hashtags or links, tweets auto-generated by applications, etc.). Three professional English speakers participated in evaluating and decising the useful tweets. The DFSMD dataset \cite{abaalkhail2018affectional} is publicly available upon request 
 
 The data were divided into three subsets, each consisting of 5000 sentences. For each dialect, an individual annotator was asked to translate only one subset (i.e., 5000 sentences) from English into his or her native Arabic dialect. For the Levantine/Shami dialect, there was one subset of 5000 sentences that had two translators to work on, each comprising 2500 sentences. The duration of the translation was approximately 6 months. Our strategy for dividing the data into small subsets and providing the translators with a long period of time was intended to ensure that they felt comfortable while doing this arduous task and would hence perform with a high level of concentration, carefully understanding the sentences, and provide consistent and accurate translations. We had 15000 translated sentences for each dialect, for a total of 60,000 translated sentences.

\subsection{Translation Strategy}
We adopted the content localization translation approach, where translating the texts not only conveys a near-equivalent meaning but also addresses and integrates linguistic, cultural, tone, and contextual components of the texts. The same words might convey different meanings in different dialects; for example, the word "chips" refers to fried thin potato chips in North American English, while it refers to "fries" in UK English. 
The same applies for different dialects in the Arabic language. For example, the word "\bgroup\beginR\raggedleft\fontencoding{LAE}\selectfont صاحبي\endR\egroup "
in Gulf dialect refers to a friend, while in Lebanese dialect, it means "boyfriend".

Since the dataset was collected from social media (i.e., Twitter in this work) and the tweets were mostly expressed in a day-to-day spoken language, the translation from English to multi dialectal Arabic was customized to cultural language of OSN. Thus, we took into consideration a number of additional criteria (i.e., in addition to the content localization-based translation approach): 

\begin{enumerate}
\item consideration of the OSN's cultural language and expressions: iconic emotion (e.g., emoticons, emojis) and hashtag words were kept in the translated texts while preserving their occurrence order and their context. Hashtag words were translated into the corresponding Arabic dialect.
\item Consideration of idiomatic expressions: an idiomatic expression should not be word-to-word translated. Instead, it should be translated to convey the context or its equivalent idiomatic expression in the corresponding dialect while preserving the context of the original text. Current translation tools were found to provide inaccurate translations of idiomatic expressions from English to Arabic language \cite{musaad2023translation}.
\item Language code borrowing was considered: Code borrowing refers to using one primary language but mixing in words from another language to fit the primary language. For example, the word "lol" is used and written using the Arabic alphabet as "\bgroup\beginR\raggedleft\fontencoding{LAE}\selectfont لول\endR\egroup"; similarly, the word "pistachio" is used and written using the Arabic alphabet as "\bgroup\beginR\raggedleft\fontencoding{LAE}\selectfont بيستاشيو\endR\egroup"

In this work, four main Arabic dialects were studied: Gulf, Yemeni, Iraqi, and Levantine/Shami. The qualified translators were provided with a subset of English sentences and translation guidelines that covered the above criteria. In addition, they were advised to convert mainly proper nouns into Arabic letters where applicable – for example, names of people ( "John" to "\bgroup\beginR\raggedleft\fontencoding{LAE}\selectfont جون\endR\egroup"), and names of places ("Lebanon" to "\bgroup\beginR\raggedleft\fontencoding{LAE}\selectfont لبنان\endR\egroup "). The translators were also advised to pay attention to the spelling, as any misspelling would harm the quality of the translation. Upon the completion of the translation task, the translators were asked to perform a round of proofreading before they submitted the final translations. Note that the translation was performed for each dialect individually with the corresponding dialect translators.

\end{enumerate}

\begin{figure*}[th]
\centering
\includegraphics[width=0.5\textwidth]{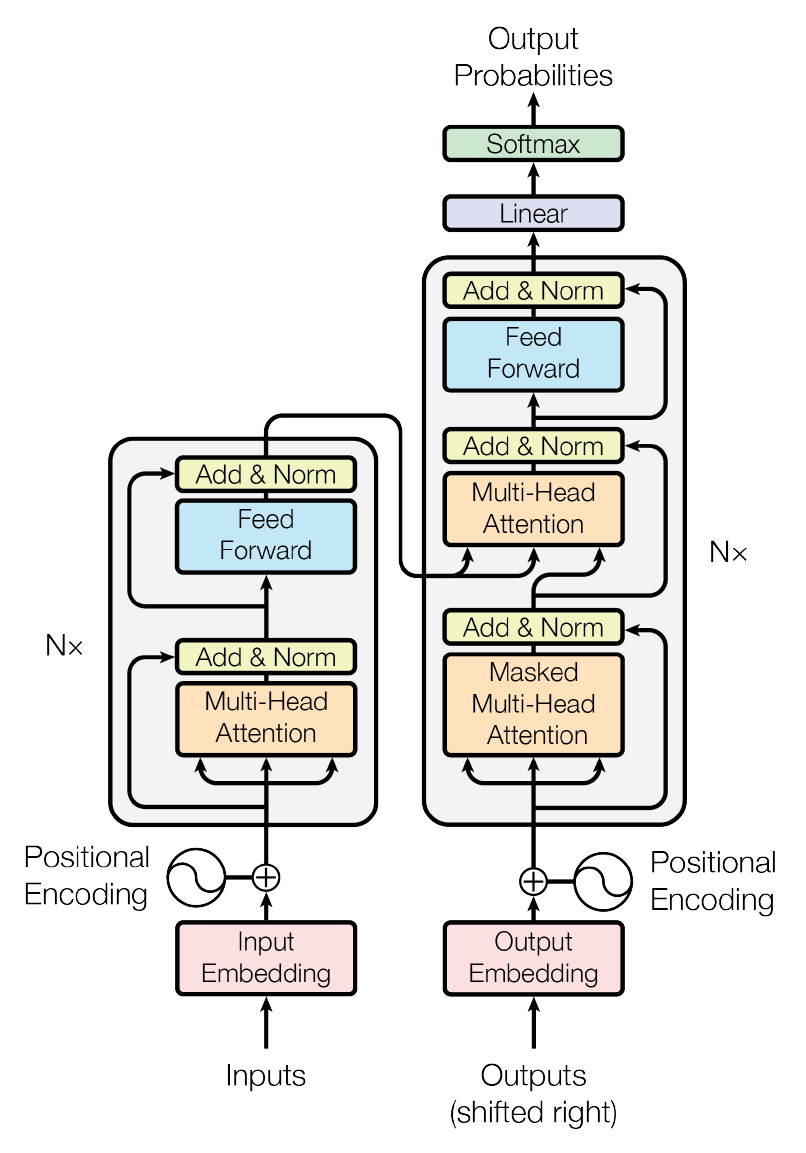}
\caption{Sequence-to-sequence Transformer architecture \cite{transformers} used in training our English-dialectal Arabic.}

\label{fig:transformers-arch}
\end{figure*}

\subsection{Translation Modeling Method}

\subsubsection{preprocessing}

\begin{itemize}

\item {Removing extra whitespaces}
\item  {Removing encoding symbols}
\item  {Removing user mentions}
\item  {Removing URLs}
\item  {Converting text to lower case}: this was applied to English words existing in the data.
\item  {Removing special characters and numbers}
\item  {Removing tashkeel and harakat}: tashkeel or harakat refer to all the diacritics placed over or below letters.
\item  {Normalizing Hamza}: Normalizing Hamza forms into one form. For example: "\bgroup\beginR\raggedleft\fontencoding{LAE}\selectfont أهلا\endR\egroup " becomes "\bgroup\beginR\raggedleft\fontencoding{LAE}\selectfont ءهلا\endR\egroup "

\end{itemize}

\subsubsection{Sequence-to-Sequence Transformer Model}

Transfer learning offers a rich set of benefits, including improving the efficiency of model training and saving time and resources, since building a high-performance model from scratch requires a large amount of data, time, resources, and effort. Therefore, we used the fine-tuning learning approach \cite{alzamzami2021monitoring, wang2021deep} to train Transformer-based machine translation models as an attempt to overcome the limitations \cite{alzamzami2020light} of small datasets and domain adaptation.

\begin{figure*}[th]
\centering
\includegraphics[width=.5\textwidth]{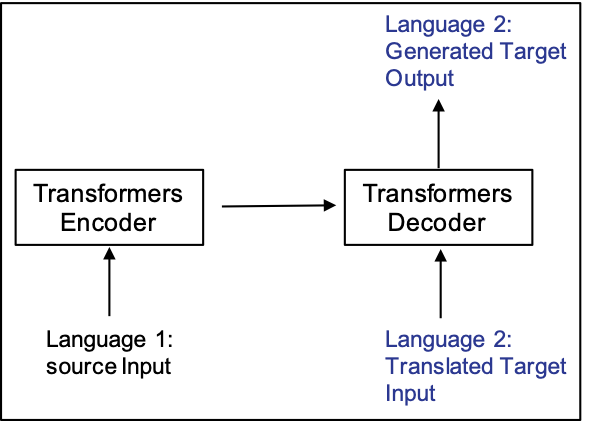}
\caption{A flow diagram of fine-tuning our neural machine translation (NMT).}
\label{fig:flow-diagram-nmt}
\end{figure*}

We used the sequence-to-sequence Transformers architecture (Figure. \ref{fig:transformers-arch}) proposed in \cite{liu2020multilingual}.  This architecture consists of 12 layers of encoder and 12 layers of decoders with a model dimension of 1024 on 16 heads. On top of the both encoder and decoder, there is an additional normalization layer that was found to stabilize the training. We used pretrained weights from the mBART \cite{liu2020multilingual} model that was pre-trained on 25 languages including the Arabic language with the corpus size of 28.0 GB. For our Arabic dialect models, we used our proposed multidialectal datasets (OSN-MDAD) to fine tune the mBART \cite{liu2020multilingual} pretrained model. The model learns parallel translation from informal and slang English to informal and dialectal Arabic. We produced four models corresponding to four dialects: Iraqi, Yemeni,Gulf,and Levantine/Shami. Figure. \ref{fig:flow-diagram-nmt} illustrates the flow diagram followed in training our neural machine translation models.

\section{Experimental Results and Analysis}
\label{sec:exprResult}

\subsection{Experiment Design and Evaluation Metric}

For evaluation purposes, we trained sequence-to-sequence (seq2seq) with attention models using our proposed dataset, with one mode for each of four dialects. We then compared the performance of the seq2seq models with our proposed Transformer-based models. 

For the evaluation metrics, we used BLUE and ROUGE metrics. BLUE metric is recommended for the translation tasks as it conducts a robust assessment of  the quality of translation fairly quickly. ROUGE complements BLUE in terms of evaluation, where the former focuses on recall while the latter is precision oriented. Therefore, we used the F-score of BLUE and ROUGE. 

The data were split with ratio of a 90\% for training and 10\% for testing. The default settings of mBART were adopted in model training.

\subsection{Performance of Transformers models}

\begin{table*}[h]
  \centering
  \large
  \caption{Comparison results between sequence-to-sequence with attention (from scratch) and sequence-to-sequence Transformer models (with transfer learning) using our proposed Arabic multidialect dataset (OSN-MDAD). The performance was evaluated using F-score of BLUE and ROUGE metrics.}
    \begin{tabular}{l|c|c|}
    \toprule
          & seq-to-seq with attention & seq-to-seq Transformers  \\
    \midrule
    Yemeni & 11.3 & \textbf{30} \\
    \midrule
    Iraqi & 1.7 & \textbf{41.7} \\
    \midrule
    Gulf & 13.1 & \textbf{33.8} \\
    \midrule
    Levantine/Shami & 1.7 & \textbf{34.6} \\
    \bottomrule
    \end{tabular}%
  \label{tab:seq2seqnmt}%
\end{table*}%

Table. \ref{tab:seq2seqnmt} shows that Transformer-based models outperform
seq-to-seq with attention models that were trained without the transfer learning approach. Transformer models that were trained by fine tuning a pretrained model (i.e., mBART in this work) yielded by far better translations, with an improvement of  $\approx$ 19 more points for Yemeni, 40 more points for Iraqi, $\approx$ 21 more points for Gulf, and $\approx$ 33 more points for Levantine/Shami. 

The nature of Transformers that uses multihead self-attention allows the models to learn better word contextualization and hence yield better results in terms of F-score of BLUE and ROUGE. Self-attention enables contextualizing every word in various positions with respect to the whole sequence. This solves the problem that similar words might have different meanings in different sequences. This problem was shown in the performance of seq-to-seq with attention models (Table. \ref{tab:seq2seqnmt}), which use attention to connect the recurrent states of the encoder to the recurrent states of the decoder.

\section{Conclusion and Future Work}
\label{sec:conclusion}

Without a doubt, the informal and slang nature of conversations has become the dominant mode for communication on social media. Moreover, a variety of languages other than English are used in social talks and conversations. Accordingly, monitoring systems for online social behavior have to extend their ability to understand and analyze interactions expressed in different languages and even dialects. In response, we propose a framework that targets low-resource language (i.e., Arabic in this work). Considering the translation approach used to solve the multilingual challenges, we propose a translation dataset (OSN-MDAD) from English to multidialectal Arabic in an effort to contribute to the low-resource status \cite{shi2022low} of multidialectal Arabic resources. Four different Arabic dialects were considered in this work: Gulf, Yemeni, Iraqi, and Levantine/Shami.
OSN-MDAD dataset was then used for the purpose of building translation models -one for each dialect- from English to multidialectal Arabic optimized for OSN data. 
Our experimental results have illustrated the superior performance of our English-dialectal Arabic translations models trained using Transformer architecture -with transfer learning- compared to the performance yielded by the sequence-to-sequence with attention mechanism -from scratch-, using our proposed translation dataset. We believe that this study contributes to the research community by providing a high-quality benchmark Arabic multidialectal translation dataset that can be used by researchers to learn from and benchmark against and also extend upon in future related tasks. Another contribution of this work is the proposed guideline framework for constructing contextual translation datasets. Some of the instructions were customized to fit the purpose of this study; however, they could be customized to fit other problems of interest. The proposed guidelines were carefully studied, developed, and considered after rounds of expert consultations.

For future directions, an extension of additional Arabic dialects will be considered to expand the ability of systems to understand and analyze broader online Arabic content. This extension will assist in enriching the under-resourced status of the Arabic language in order to benefit many downstream tasks. Additionally, we are interested in investigating long sentence translation in the context of multiple Arabic dialects, as this work has only experimented with short texts. The benefits of translation systems that are able to handle both short and long sentences are great and will significantly reduce the expensive cost of building individual models for each case. While the BLUE evaluation metric provides the average performance measure of how well a machine translation model translates content, it is limited in providing insights into where the model struggles to produce fluent translations. Therefore, we plan to explore the evaluation area in order to study aspects such as linguistics to enhance the performance of machine translation models.

\bibliographystyle{unsrt}  
\bibliography{references}  

\end{document}